\documentclass[10pt,twocolumn,letterpaper]{article}

\usepackage{iccv}
\usepackage{times}
\usepackage{epsfig}
\usepackage{graphicx}
\usepackage{amsmath}
\usepackage{amssymb}

\usepackage{kotex}

\usepackage{adjustbox}
\usepackage{booktabs}
\usepackage[table, dvipsnames]{xcolor}
\usepackage{color, colortbl}
\usepackage{multirow}

\renewcommand{\etal}{\textit{et al.}}
\newcommand{\myparagraph}[1]{\vspace{4pt}\noindent{\bf #1}}

\usepackage[pagebackref=true,breaklinks=true,colorlinks,bookmarks=false]{hyperref}

\iccvfinalcopy 

\ificcvfinal\fi

\begin{document}

\title{SelfReg: Self-supervised Contrastive Regularization for Domain Generalization}

\author{Daehee Kim$^1$, Seunghyun Park$^2$, Jinkyu Kim$^3$, and Jaekoo Lee$^1$ \\
\small{$^1$ College of Computer Science, Kookmin University}\\
\small{$^2$Clova AI Research, NAVER Corp.} \\
\small{$^3$ Department of Computer Science and Engineering, Korea University}
}


\maketitle

\begin{abstract}
In general, an experimental environment for deep learning assumes that the training and the test dataset are sampled from the same distribution. However, in real-world situations, a difference in the distribution between two datasets, domain shift, may occur, which becomes a major factor impeding the generalization performance of the model. The research field to solve this problem is called domain generalization, and it alleviates the domain shift problem by extracting domain-invariant features explicitly or implicitly.
In recent studies, contrastive learning-based domain generalization approaches have been proposed and achieved high performance. These approaches require sampling of the negative data pair. However, the performance of contrastive learning fundamentally depends on quality and quantity of negative data pairs.
To address this issue, we propose a new regularization method for domain generalization based on contrastive learning, self-supervised contrastive regularization (SelfReg). The proposed approach use only positive data pairs, thus it resolves various problems caused by negative pair sampling. Moreover, we propose a class-specific domain perturbation layer (CDPL), which makes it possible to effectively apply mixup augmentation even when only positive data pairs are used.
The experimental results show that the techniques incorporated by SelfReg contributed to the performance in a compatible manner. In the recent benchmark, DomainBed, the proposed method shows comparable performance to the conventional state-of-the-art alternatives. 
Codes are available at \href{https://github.com/dnap512/SelfReg}{https://github.com/dnap512/SelfReg}.
\end{abstract}

\section{Introduction}
Machine learning systems often fail to generalize out-of-sample distribution as they assume that in-samples and out-of-samples are independent and identically distributed -- this assumption rarely holds during deployment in real-world scenarios where the data is highly likely to change over time and space. Deep convolutional neural network features are often domain-invariant to low-level visual cues~\cite{peng2015learning}, 
some studies~\cite{donahuedeep} suggest that they are still susceptible to domain shift.

\begin{figure}[t]
\begin{center}
   \includegraphics[width=\linewidth]{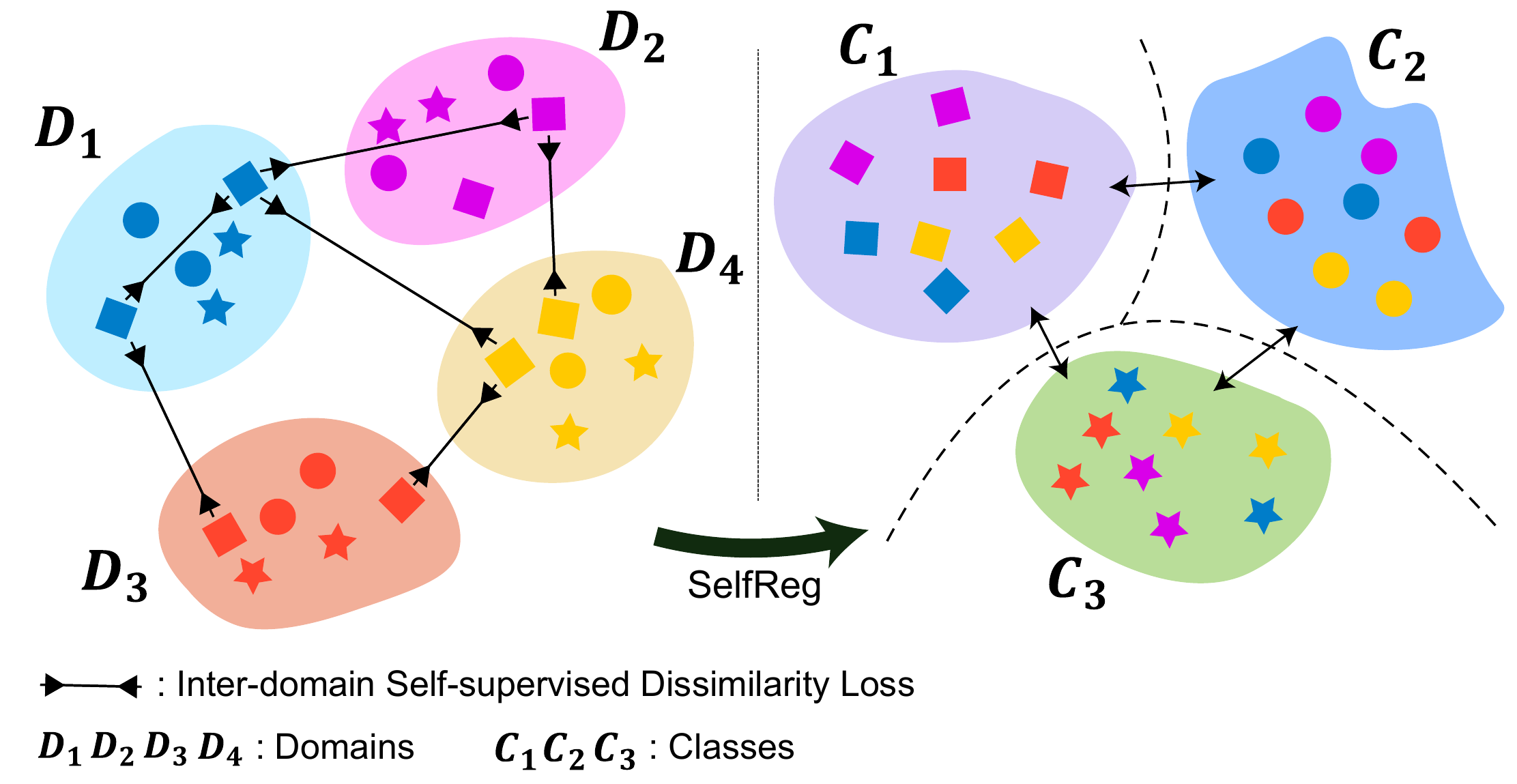}
\end{center}
   \caption{Our model utilizes the self-supervised contrastive losses for the model to learn domain-invariant representation by mapping the latent representation of the same-class samples close together. Note that different shapes (i.e. circles, stars, and squares) indicate different classes $\mathcal{C}_{i\in\{1,2,3\}}$, and we differently color-code according to their domain $\mathcal{D}_{i\in\{1,2,3,4\}}$.}
\label{fig:teaser} 
\end{figure}

There have been increasing efforts to develop models that can generalize well to out-of-distribution. The literature in domain generalization (DG) aims to learn the invariances across multiple different domains so that a classifier can robustly leverage such invariances in unseen test domains~\cite{vapnik1998statistical, ganin2016domain, li2018deep, li2018domain, muandet2013domain, sun2016deep}. In the domain generalization task, it is assumed that multiple source domains are accessible during training, but the target domains are not~\cite{blanchard2011generalizing, muandet2013domain}. This is different from domain adaptation (DA), semi-supervised domain adaptation (SSDA), and unsupervised domain adaptation (UDA) problems, where examples from the target domain are available during training. In this paper, we focus on the domain generalization task.

Some recent studies~\cite{chen2020simple,he2020momentum,oord2018representation} suggest that contrastive learning can be successfully used in a self-supervised learning task by mapping the latent representations of the positive pair samples close together, while that of negative pair samples further away in the embedding space. Such a contrastive learning strategy has also been utilized for the domain generalization tasks~\cite{motiian2017unified, dou2019domain}, similarly aiming to reduce the distance of same-class features in the embedding space, while increasing the distance of different-class features. However, such negative pairs often make the training unstable unless useful negative samples are available in the same batch, which is but often challenging. 

In this work, we revisit contrastive learning for the domain generalization task, but only with positive pair samples, as shown in Figure~\ref{fig:teaser}. As it is generally known that using positive pair samples only causes the performance drop, which is often called representation collapse~\cite{grill2020bootstrap}. Inspired by recent studies on self-supervised learning~\cite{chen2020exploring, grill2020bootstrap}, which successfully avoids representation collapse by placing one more projection layer at the end of the network, we successfully learn domain-invariant features and our model trained with self-supervised contrastive losses shows the matched or better performance against alternative state-of-the-art methods, where ours is ranked at 2nd places in the domain generalization benchmarks, i.e. DomainBed~\cite{gulrajani2020search}. 

However, self-supervised contrastive losses are only part of the story. As we generally use a linear form of the loss function, properly balancing gradients is required so that network parameters converge to generate domain-invariant features. To mitigate this issue, we advocate for applying the following three gradient stabilization techniques: (i) loss clipping, (ii) stochastic weights averaging (SWA), and (iii) inter-domain curriculum learning (IDCL). We observe that the combined use of these techniques further improves the model's generalization power. 

To effectively evaluate our proposed model, we first use the publicly available domain generalization data set called PACS~\cite{Li2017dg}, where we analyzed our model in detail to support our claims. We further experiment with much larger benchmarks called DomainBed~\cite{gulrajani2020search} where our model shows matched or better performance against alternative state-of-the-art methods.

We summarize our main contributions as follows:
\begin{itemize}
    \item SelfReg facilitates the application of metric learning using only positive pairs without negative pairs. \vspace{-0.5em}
    \item We devised a CDPL by exploiting a condition that use only positive pairs. The combination of CDPL and mixup improves the weakness of mixup approach. \vspace{-0.5em}
    \item The performance comparable to that of the SOTA DG methods was confirmed in the DomainBed that facilitated the comparison of DG performance in the fair and realistic environment. \vspace{-0.5em}
\end{itemize}

\begin{figure*}[t]
\begin{center}
  \includegraphics[width=\linewidth]{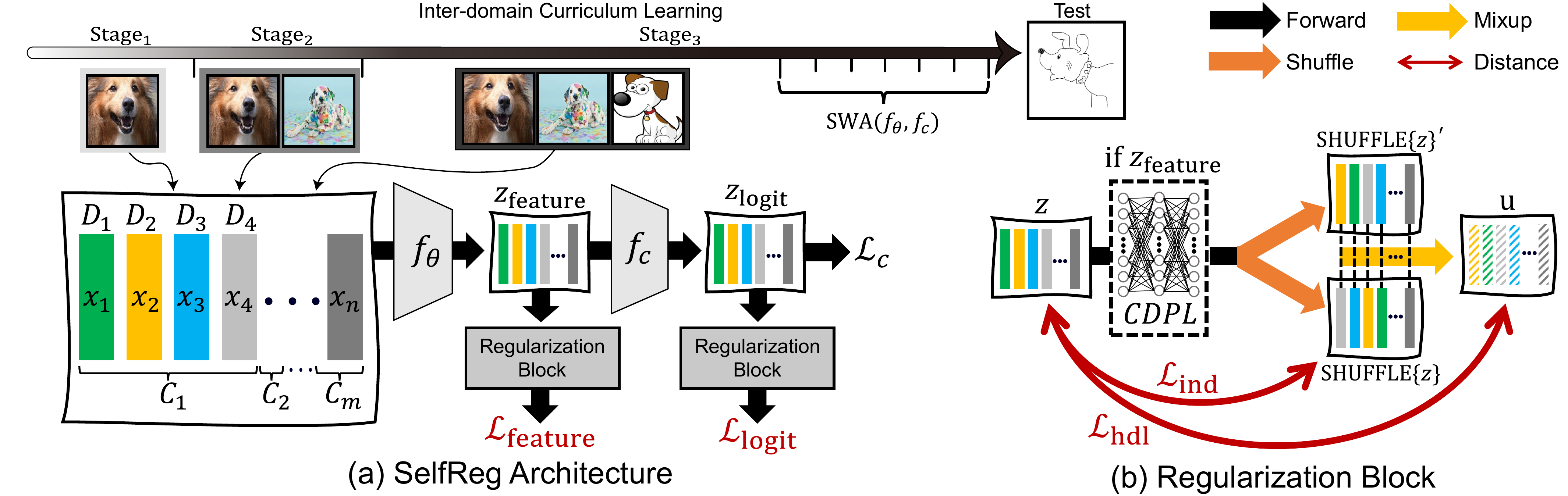}
\end{center}
  \caption{An overview of our proposed SelfReg. Here, we propose to use the self-supervised (in-batch) contrastive losses to regularize the model to learn domain-invariant representations. These losses regularize the model to map the representations of the ``same-class'' samples close together in the embedding space. We compute the following two dissimilarities in the embedding space: (i) individualized and (ii) heterogenerous self-supervised dissimilarity losses. We further use the stochastic weight average (SWA) technique and the inter-domain curriculum learning (IDCL) to optimize gradients in conflict directions.}
\label{fig:method}
\end{figure*}

\section{Related Work}
The main goal of domain generalization (DG) is to generate domain-invariant features so that the model is generalizable to unseen target domains, which are generally outside the training distribution. Of a landmark work, Vapnik~\etal~\cite{vapnik1998statistical} introduces Empirical Risk Minimization (ERM) that minimizes the sum of errors across domains. Notable variants have been introduced to learn domain-invariant features by matching distributions across different domains. Ganin~\etal~\cite{ganin2016domain} utilizes an adversarial network to match such distributions, while Li~\etal~\cite{li2018deep} instead matches the conditional distributions across domains. Such a shared feature space is optimized by minimizing maximum mean discrepancy~\cite{li2018domain}, transformed feature distribution distance~\cite{muandet2013domain}, or covariances~\cite{sun2016deep}. In this work, we also follow this stream of work, but we explore the benefit of self-supervised contrastive learning that can inherently learn to domain-invariant discriminating feature by explicitly mapping the ``same-class'' latent representations close together. 

To our best knowledge, there are few that applied contrastive learning in the domain generalization setting. Classification and contrastive semantic alignment (CCSA)~\cite{motiian2017unified} and model-agnostic learning of semantic features (MASF)~\cite{dou2019domain} aimed to reduce the distance of same-class (positive pair) feature distributions while increasing the distance of different-class (negative pair) feature distributions. However, using such negative pairs often make the training unstable unless useful negative samples are available in the same batch, which is often challenging. To address this issue, we focus on minimizing a distance between the same-class (positive pair) features in the embedding space as recently studied for the self-supervised learning task~\cite{chen2020simple,he2020momentum,oord2018representation}, including BYOL~\cite{grill2020bootstrap} and SimSiam~\cite{chen2020exploring}.

Inter-domain mixup~\cite{yan2020improve, xu2020adversarial, wang2020heterogeneous} techniques are introduced to perform empirical risk minimization on linearly interpolated examples from random pairs across domains. We also utilize such a mixup, but we only interpolate same-class features to preserve the class-specific features. We observe that such a same-class mixup help obtaining robust performance for unseen domain data.

As another branch, JiGen~\cite{carlucci2019domain} utilizes a self-supervised signal by solving a jigsaw puzzle as a secondary task to improve generalization. Meta-learning frameworks~\cite{li2018learning} are also explored for domain generalization to meta-learn how to generalize across domains by leveraging MAML~\cite{finn2017model}. Some also explored splitting the model into domain-invariant and domain-variant components by low-rank parameterization~\cite{Li2017dg}, style-agnostic network~\cite{nam2019reducing}, domain-specific aggregation modules~\cite{d2018domain}.

\section{Method}\label{sec:method}
We start by motivating our method before explaining its details. The main goal of domain generalization is to learn a domain-invariant representation from multiple source domains so that a model can generalize well across unseen target domains. While domain-variant representation can be achieved to some degree through deep network architectures, invariant representations are often harder to achieve and are usually implicitly learned with the task. To address this, we argue that a model should learn a domain-invariant discriminating feature by comparing among different samples -- the comparison can be performed between positive pairs of same-class inputs and negative pairs of different-class inputs. 

Here we propose the self-supervised contrastive losses to regularize the model to learn domain-invariant representation by mapping the representations of the ``same-class'' samples close together, while that of ``different-class'' samples further away in the embedding space. This may share a similar idea with contrastive learning, which trains a discriminative model on multiple input pairs according to some notion of similarity. Thus, we start with the recent batch contrastive approaches and extend them to the domain generalization setting, as shown in Figure~\ref{fig:method}. While some domain generalization approaches need to modify the model architecture during learning, our proposed contrastive method is much simpler where no modification to the model architecture is needed.

In the next section, we explain our proposed self-supervised contrastive losses for domain generalization tasks, which mainly measures the following two feature-level dissimilarities in the embedding space: (i) {\em Individualized In-batch Dissimilarity Loss} (Section~\ref{ss:pdl}) and (ii) {\em Heterogeneous In-batch Dissimilarity Loss} (Section~\ref{ss:hdl}). Note that these losses can be applied to both the intermediate features and the logits from the classifier (Section~\ref{ss:losses}). In fact, in our ablation study (Section~\ref{ss:ablation_study}), the combined use of both regularization achieves the best performance. In Section~\ref{ss:swa}, we also discuss the stochastic weight average (SWA) technique that we use with our self-supervised contrastive losses and observe a further performance improvement, which is possibly due to SWA provides the more flatness in loss surface by ensembling domain-specific models. 
We exclude contents of the inter-domain curriculum learning (IDCL) strategy in this paper because of publication copyright issues.

\subsection{Individualized In-batch Dissimilarity Loss}~\label{ss:pdl}
Given latent representations ${\bf{z}}_{i}^{c}=f_{\theta}({\bf{x}}_{i})$ for $i \in \{1, 2, \dots, N\}$ and a class label $c\in\mathcal{C}$, we compute the individualized in-batch dissimilarity loss $\mathcal{L}_\textnormal{ind}$. Note that we use a feature generator $f_{\theta}$ parameterized by $\theta$ and we use a batch size of $N$. The dissimilarity between a positive pair of the ``same-class'' latent representations is measured as in the following Eq.~\ref{eq:L_ind}:
\begin{equation}\label{eq:L_ind}
    \mathcal{L}_\textnormal{ind}({\bf z}) = \frac{1}{N}\sum_{i=1}^{N}\big|\big|{\bf{z}}^{c}_{i}-f_{\textnormal{CDPL}}\big({\bf{z}}^{c}_{j\in[1,N]}\big)\big|\big|^{2}_{2}
\end{equation}
where ${\bf{z}}^{c}_{j}$ is randomly chosen from other in-batch latent representations $\{{\bf{z}}^{c}_{i}\}$ that has the same class label $c\in\mathcal{C}$. Note that we only consider optimizing the alignment of positive pairs and the uniformity of the representation distribution at the same time. As discussed in \cite{grill2020bootstrap}, we use an additional an MLP layer $f_{\textnormal{CDPL}}$, called Class-specific Domain Perturbation Layer, to prevent the performance drop caused by so-called representation collapse. We provide an ablation study in Section~\ref{ss:ablation_study} to confirm the use of $f_{\textnormal{CDPL}}$ achieves better performance.  

\begin{figure}[t]
\begin{center}
   \includegraphics[width=\linewidth]{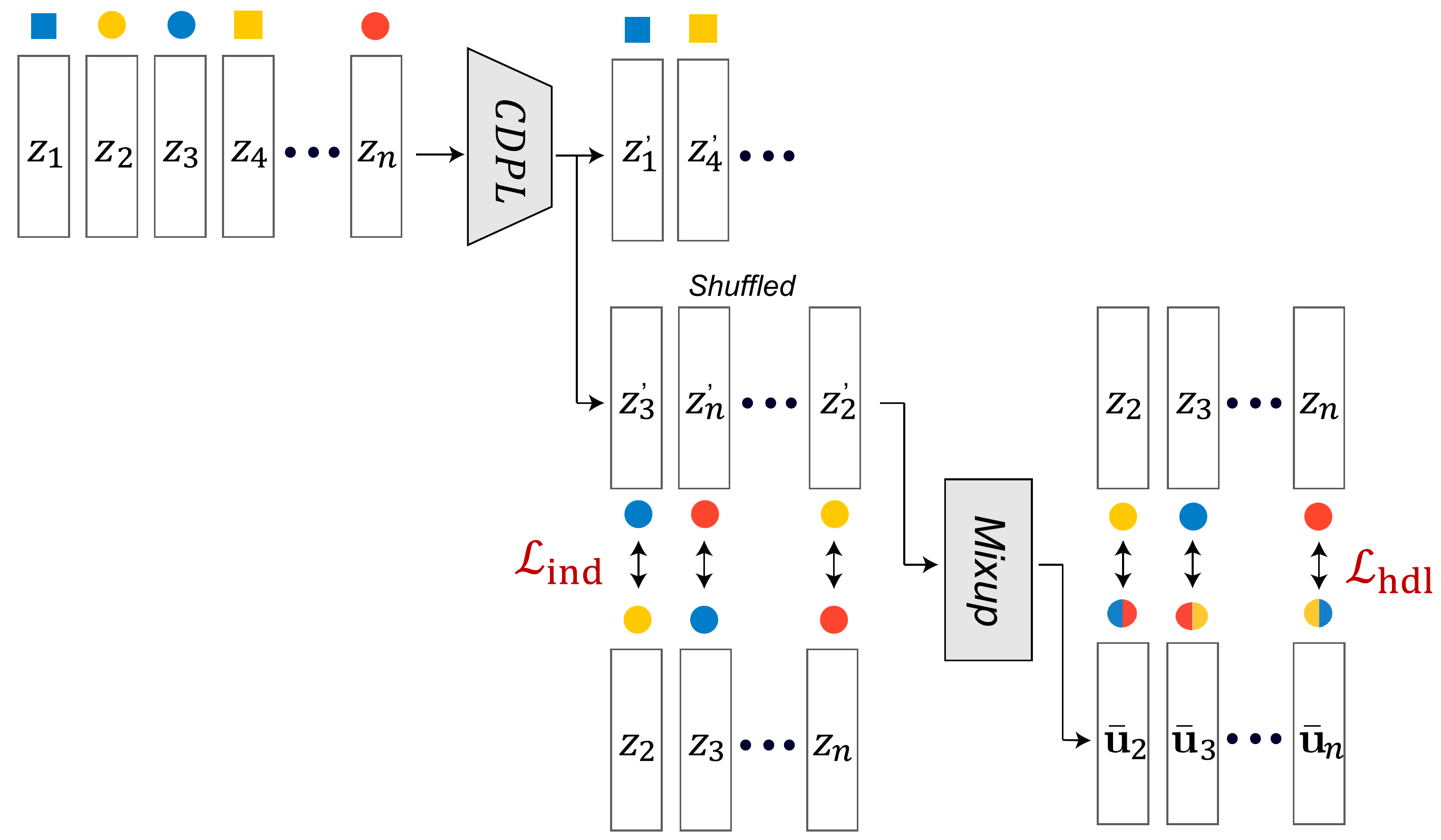}
\end{center}
   \caption{An overview of our proposed self-supervised contrastive regularization losses.}
\label{fig:losses}
\end{figure}

For better computational efficiency, we use the following two steps to find all positive pairs, as shown in Figure~\ref{fig:losses}. (i) We first cluster and order latent representations ${\bf{z}}_{i}$ into a same-class group, i.e. $\{{\bf{z}}^{c}_{i}\}$ for $c\in\mathcal{C}$. (ii) For each same-class group, we modify its order by random shuffling and obtain $\texttt{SHUFFLE}\{{\bf{z}}^{c}_{i}\}$. (iii) We finally form a positive pair in order from $\{{\bf{z}}^{c}_{i}\}$ and $\texttt{SHUFFLE}\{{\bf{z}}^{c}_{i}\}$. 

\subsection{Heterogeneous In-batch Dissimilarity Loss}~\label{ss:hdl}
To further push the model to learn domain-invariant representations, we use an additional loss, called heterogeneous in-batch dissimilarity loss. Given latent representations ${\bf{u}}_{i}=f_{\textnormal{CDPL}}({\bf{z}}^{c}_{i})$ from the previous step, we apply a two-domain Mix-up layer to obtain the interpolated latent representation ${\bar{\bf{z}}}_{i}$ across different domains. This regularizes the model on the mixup distribution~\cite{zhang2017mixup}, i.e. a convex combination of samples from different domains. This is similar to a layer proposed by Wang~\etal~\cite{wang2020heterogeneous} as defined as follows:
\begin{equation}
    {\bar{\bf{u}}}^{c}_{i} = \gamma{\bf{u}}^{c}_{i} + (1-\gamma){\bf{u}}^{c}_{j\in[1,N]} 
\end{equation}
where $\gamma\sim\texttt{Beta}(\alpha, \beta)$ for $\alpha = \beta \in (0, \infty)$. Similarly, ${\bf{u}}^{c}_{j}$ is randomly chosen from $\{{\bf{u}}^{c}_{i}\}$ for $i\in\{1,2,\dots,N\}$ that have the same class label. Note that $\gamma\in[0,1]$ is controlled by hyper-parameters $\alpha$ and $\beta$.

Finally, we compute the heterogeneous in-batch dissimilarity loss $\mathcal{L}_\textnormal{hdl}({\bf z})$ as follows:
\begin{equation}\label{eq:L_hdl}
    \mathcal{L}_\textnormal{hdl}({\bf z}) = \frac{1}{N}\sum_{i=1}^{N}\big|\big|{\bf{z}}^{c}_{i}-{\bar{\bf{u}}}^{c}_{i}\big|\big|^{2}_{2}
\end{equation}

\subsection{Feature and Logit-level Self-supervised Contrastive Losses}~\label{ss:losses}
The proposed individualized and heterogeneous in-batch dissimilarity losses can be applied to both the intermediate features and the logits from the classifier. We use the loss function $\mathcal{L}_\textnormal{SelfReg}$ as follows:
\begin{equation}
    \mathcal{L}_\textnormal{SelfReg} = \lambda_\textnormal{feature}\mathcal{L}_\textnormal{feature} + \lambda_\textnormal{logit}\mathcal{L}_\textnormal{logit}
\end{equation}
where we use $\lambda_\textnormal{feature}$ and $\lambda_\textnormal{logit}$ to control the strength of each term. As we use a linear form of the loss function, which often needs to be properly balanced so that network parameters converge to generate domain-invariant features that are also useful for the original classification task. We observe that our self-supervised contrastive losses $\mathcal{L}_\textnormal{SelfReg}$ become dominant after the initial training stage, inducing gradient imbalances to impede proper training. 
To mitigate this issue, we apply two gradient stabilization techniques: (i) loss clipping and (ii) stochastic weights averaging (SWA), and (iii) inter-domain curriculum learning (IDCL). For (i), we modify gradient magnitudes to be dependent on the magnitude of the classification loss $\mathcal{L}_\textnormal{c}$ -- i.e. we use the gradient magnitude modifier $\texttt{min}(1.0, \mathcal{L}_\textnormal{c})$ and thus $\mathcal{L}_\textnormal{feature} = \texttt{min}(1.0, \mathcal{L}_\textnormal{c})\big[\gamma\mathcal{L}_\textnormal{ind} + (1-\gamma)\mathcal{L}_\textnormal{hdl}\big]$. For (ii) we discuss details in Section~\ref{ss:swa}.

\myparagraph{Loss Function}
Ultimately, we use the following loss function $\mathcal{L}$ that consists of classification loss $\mathcal{L}_\textnormal{c}$ as well as our self-supervised contrastive loss $\mathcal{L}_\textnormal{SelfReg}$:
\begin{equation}
    \mathcal{L} = \mathcal{L}_\textnormal{c} + \mathcal{L}_\textnormal{SelfReg}
\end{equation}

\subsection{Stochastic Weights Averaging (SWA)}\label{ss:swa}
Stochastic Weight Average~(SWA) is an ensembling technique to find a flatter minimum in loss space by averaging snapshots of model parameters derived from multiple local minima in the training procedure~\cite{izmailov2018averaging}. It is known that finding a flatter minima guarantees better generalization performance~\cite{he2019asymmetric}, and thus it has been used in domain adaptation and generalization fields that require high generalization performance~\cite{zhao2019learning, cha2021domain}.

Given model weight space $\Omega=\{\omega_0, \omega_1, \dots, \omega_N\}$, where $N$ is the number of training steps. There is no specific constraint for sampling model weights, however, in general, sampling process is performed at a specific period while the model is sufficiently converged. We use $c$ as a cyclic step length and sample weight space for SWA is $\Omega_\textnormal{swa}=\{\omega_{m+kc}\}$ for $k \geq 0, 0 \leq m \leq m+kc \leq N$, where $m$ indicates the initial step for SWA. Then we can derive the averaged weight $w_\textnormal{swa}$ as follows:
\begin{equation}
    \omega_\textnormal{swa} = \frac{1}{k+1}\sum_{i=0}^{k}{\omega_{m+ic}}. 
\end{equation}
Note that we use SWA to examine whether the proposed method is compatible with other existing techniques. Therefore, we apply SWA only for ablation study (Section~\ref{ss:ablation_study}), not DomainBed (Section~\ref{sec:domainbed}), for a fair comparison.

\begin{figure*}[t]
\begin{center}
   \includegraphics[width=\linewidth]{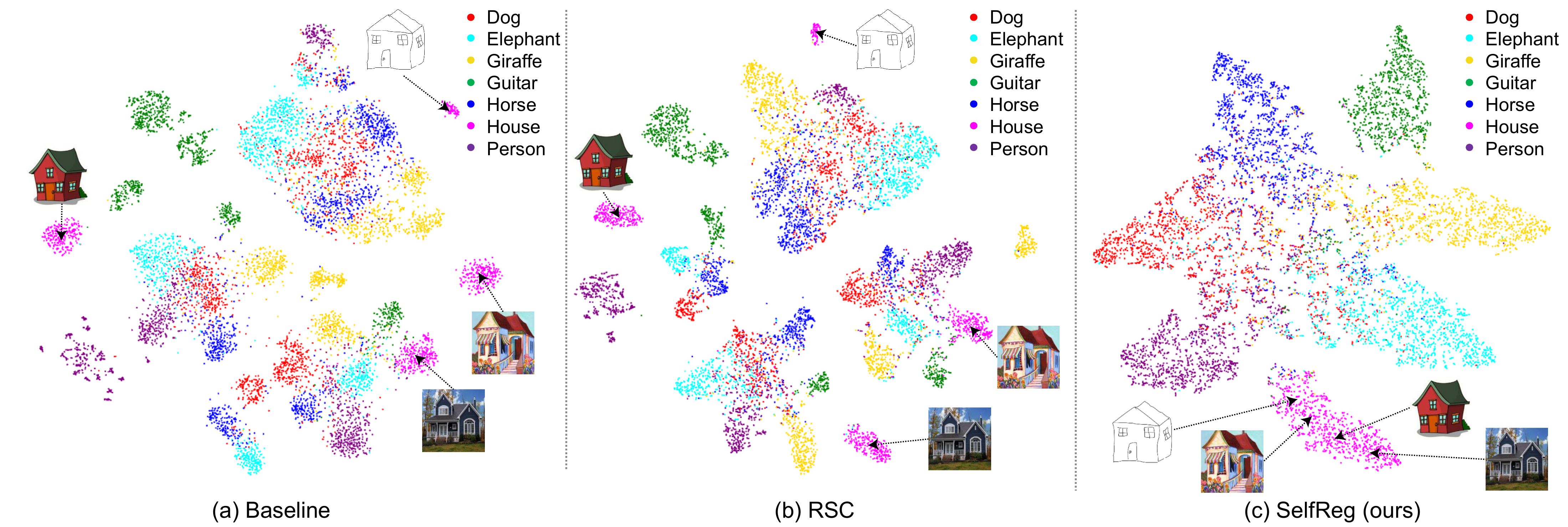}
\end{center}
   \caption{Visualizations by t-SNE~\cite{van2008visualizing} for (a) baseline (no DG techniques), (b) RSC~\cite{huangRSC2020}, and (c) ours. For better understanding, we also provide sample images of house from all target domains. Note that we differently color-coded each points according to its class. {\textit{Data}}: PACS~\cite{Li2017dg}}
\label{fig:tsne}
\end{figure*}

\section{Proof-of-Concept Experiments}
\subsection{Implementation and Evaluation Details}
Following Huang~\etal~\cite{huangRSC2020}, we train our model, for approximately 30 epochs, with a SGD optimizer using ResNet18~\cite{he2016deep} as a backbone, which is pretrained on ImageNet~\cite{deng2009imagenet}. Our backbone produces $512$-dimensional latent representation from the last layer. The batch size is set to 128 and learning rate to 0.004, which is decayed to $0.1$ at $24$ epochs. Note that such a decaying learning rate is not used when it combined with the Stochastic Weights Averaging technique, where we instead compute the averaged weight $w_\textnormal{swa}$ at the every end of each epoch. 

The loss weights are $\lambda_\textnormal{feature}=0.3$ and $\lambda_\textnormal{logit}=1.0$ were determined using grid-search. For a two-domain Mix-up layer, we use $\alpha=\beta=0.5$. The model architecture for the class-specific domain perturbation layer $f_\textnormal{CDPL}$ is a 2-layer MLPs with the number of hidden units set to $1024$, where we apply batch normalization followed by ReLU activation function. Following RSC~\cite{huangRSC2020}, data augmentation is used in our experiments to improve model generalizability. This is done by randomly cropping, flipping horizontally, jittering color, and changing the intensity.

\myparagraph{Dataset}
To verify the effectiveness of the proposed method, we evaluate our proposed method on the publicly available PACS~\cite{Li2017dg}. This benchmark dataset contains the overall 10k images from four different domains: $Photo$, $Art \ Painting$, $Cartoon$, and $Sketch$. This dataset is particularly useful in domain generalization research as it provides a bigger domain shift than existing photo-only benchmarks. This dataset provides seven object categories: i.e. dog, elephant, giraffe, guitar, horse, house, and person. We follow the same train-test split strategy from \cite{Li2017dg}, we split examples from training domains to 9:1 (train:val) and test on the whole held-out domain. Note that we use the best-performed model on validation for testing.

{
\setlength{\tabcolsep}{4pt}
\renewcommand{\arraystretch}{1.3} 
\begin{table}[t]
	\begin{center}
	    \caption{Image recognition accuracy (\%) comparison with the state-of-the-art approach, RSC~\cite{huangRSC2020}, on PACS~\cite{Li2017dg} test set. We also report standard deviation from a set of 20 models individually trained for each model and each test domain.}
	    \label{tab:pacs_benchmark}
    	\resizebox{\linewidth}{!}{%
    	\begin{tabular}{@{}lccccc@{}} \toprule
    	    \multirow{2}{*}{Model} & \multicolumn{4}{c}{Test Domain} & \multirow{2}{*}{Average} \\\cmidrule{2-5}
    	    & \textit{Photo} & \textit{Art\ Painting} & \textit{Cartoon} & \textit{Sketch} & \\ \midrule
        	A. DeepAll & 95.66 $\pm$ 0.4  &79.89 $\pm$ 1.3  &75.61 $\pm$ 1.5  &73.33 $\pm$ 2.8  &81.12 $\pm$ 0.8    \\
            B. RSC~\cite{huangRSC2020} &94.56 $\pm$ 0.4  &79.88 $\pm$ 1.7  &76.87 $\pm$ 1.2  &77.11 $\pm$ 2.7  &82.10 $\pm$ 0.9 \\\midrule
            C. A + SelfReg (ours)  &\textbf{96.22 $\pm$ 0.3}  &\textbf{82.34 $\pm$ 0.5}  &\textbf{78.43 $\pm$ 0.7} &\textbf{77.47 $\pm$ 0.8}  &\textbf{83.62 $\pm$ 0.3} \\
            \bottomrule
        \end{tabular}}
     \end{center}
\end{table}
}
\subsection{Performance Evaluation}
In Table~\ref{tab:pacs_benchmark}, we first compare our model with the state-of-the-art method, called Representation Self-Challenging (RSC)~\cite{huangRSC2020}, which iteratively discards the dominant features during training and thus encourages the network to fully use remaining features for the final verdict. For a fair comparison, all models use the identical backbone ConvNet, i.e. ResNet18. To see the performance variance, we trained each model 20 times for each test domain and report the average image recognition accuracy and its standard deviation. As shown in Table~\ref{tab:pacs_benchmark}, our proposed model clearly outperforms the other approaches in all test domains (compare the model B vs. model C), and the average image recognition accuracy is 1.52\% better than RSC~\cite{li2018domain}, while produces lower model variance (0.9 vs. 0.3 on average). 

\myparagraph{Qualitative Analysis by t-SNE}
We use t-SNE~\cite{van2008visualizing} to compute pairwise similarities in the latent space and visualize in a low dimensional space by matching the distributions by KL divergence. In Figure~\ref{fig:tsne}, we provide a comparison of t-SNE visualizations of baseline, RSC, and ours. The better a model generalizes well, the points in the t-SNE should be more clustered. As shown in Figure~\ref{fig:tsne}, (a) the baseline model and (b) RSC~\cite{huangRSC2020} produce scattered multiple clusters for each domain and class (see houses in the different clusters according to their domain). Ours is not the case for this. As shown in Figure~\ref{fig:tsne} (c), objects from the same class tend to form a merged cluster, making latent representations close to each other in the high-dimensional space.

\begin{figure}[ht]
\begin{center}
   \includegraphics[width=\linewidth]{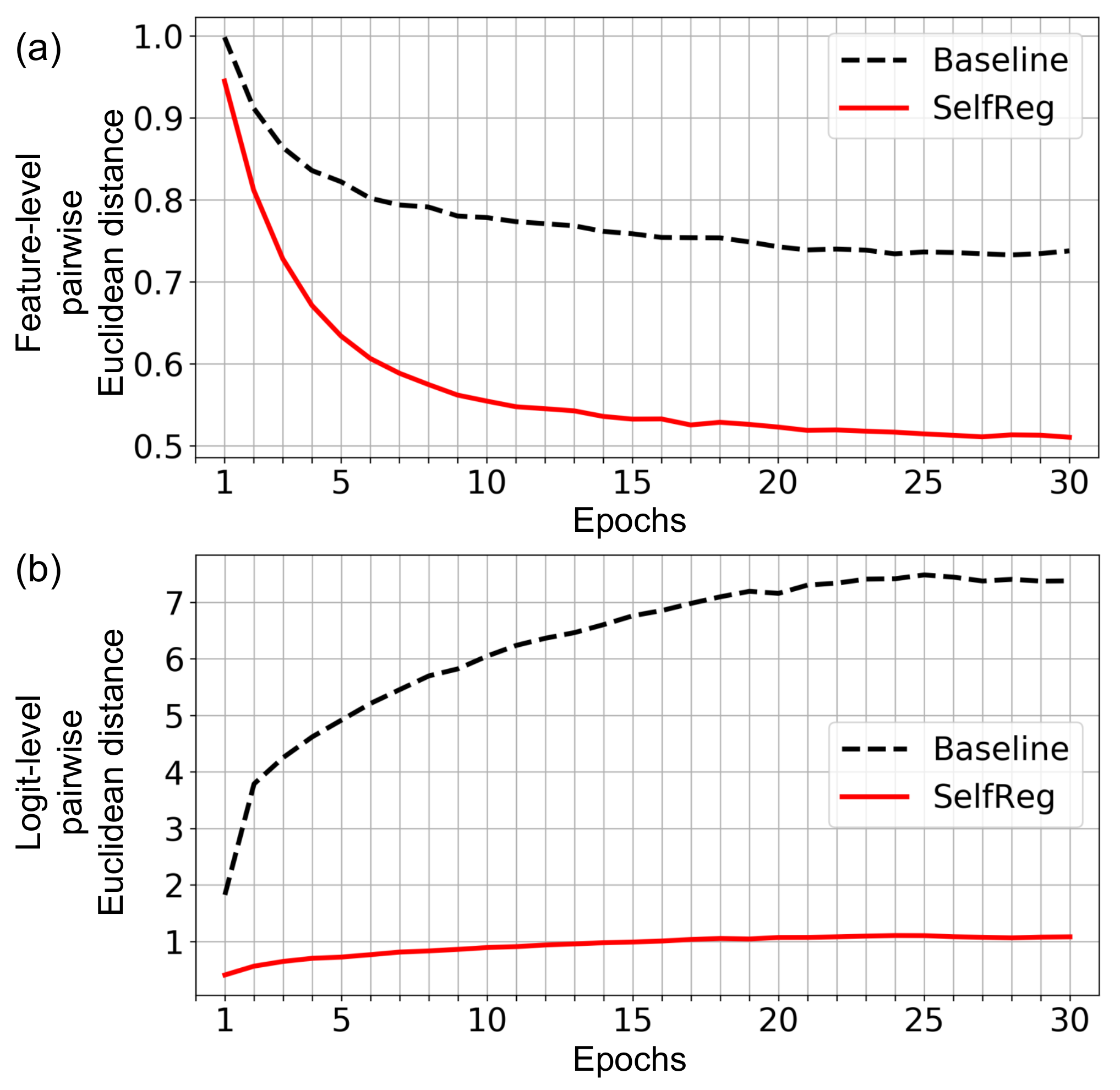}
\end{center}
   \caption{Distance between (a) a pair of same-class features and (b) a pair of same-class logits. We measure such distance at every epoch during training and compare ours (solid red) with baseline (dotted). Euclidean-based distance is used to measure distance in feature space. \textit{Data}: PACS~\cite{Li2017dg}}
\label{fig:distance}
\end{figure}
\myparagraph{The Effect of Dissimilarity Loss}
We propose two types of self-supervised contrastive loss that map the "same-class" samples close together. We observe in Figure~\ref{fig:distance} that "same-class" pairwise distance is effectively regularized in both latent (a) feature and (b) logit space (compare dotted (baseline) vs. red solid line (ours)). This was not the case for the baseline. Note that we use Euclidean-based distance to measure the pairwise difference.

\begin{figure}[ht]
\begin{center}
   \includegraphics[width=\linewidth]{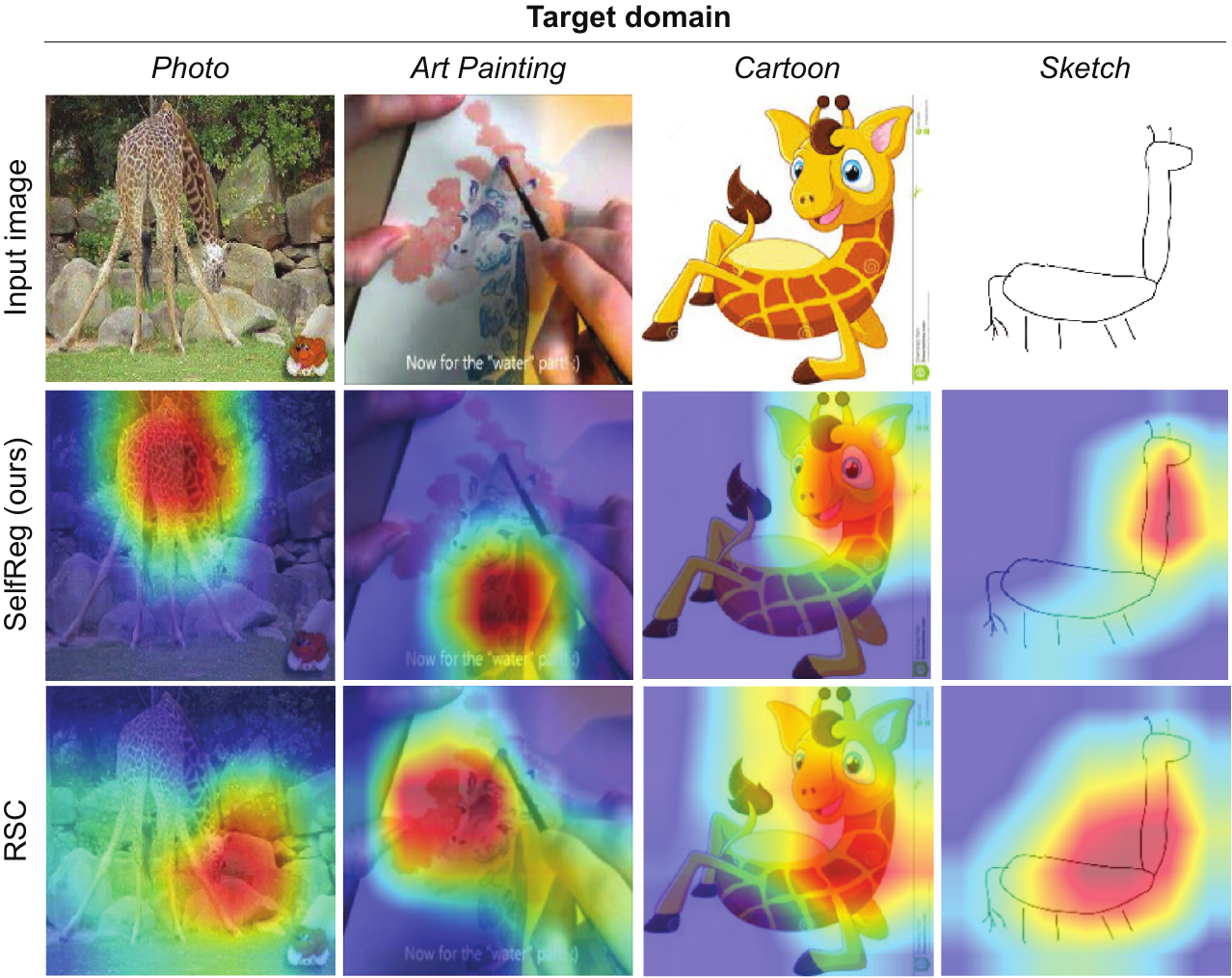}
\end{center}
   \caption{Original images with a giraffe for different domains (1st row). We provide visualizations of Grad-CAM~\cite{selvaraju2017grad} for ours and RSC~\cite{huangRSC2020} , which localizes class-discriminative regions. \textit{Data}: PACS~\cite{Li2017dg}}
\label{fig:gradcam}
\end{figure}
\myparagraph{Analysis with GradCAM}
We use GradCAM~\cite{selvaraju2017grad} to visualize image regions where the network attends to. In Fig.~\ref{fig:gradcam}, we provide examples for different target domains where we compare the model's attention maps. We observe ours better captures the class-invariant feature (i.e. the long neck of the giraffe), while RSC~\cite{huangRSC2020} does not. Red is the attended region for the network's final verdict.

{
\setlength{\tabcolsep}{4pt}
\renewcommand{\arraystretch}{1.3} 
\begin{table*}[t]
	\begin{center}
	    \caption{As an extreme case for the domain generalization task, we train our model with a single source domain (rows) and evaluate with other remaining target domains (columns). As a baseline, we also compare with RSC~\cite{huangRSC2020} of the same setting (compare left and right tables). We also report their differences in the last row ($+$ indicates that ours performs better).}
	    \label{tab:one_source_domain}
    	\resizebox{\linewidth}{!}{%
    	\begin{tabular}{@{}lccccclccccc@{}} \toprule
    	    \multirow{2}{*}{{\bf{RSC}}~\cite{huangRSC2020}} & \multicolumn{5}{c}{Target domain} & \multirow{2}{*}{\bf{SelfReg}} &\multicolumn{5}{c}{Target domain} \\ \cmidrule{2-6}\cmidrule{8-12}
            & \textit{Photo} & \textit{Art Painting} & \textit{Cartoon} & \textit{Sketch} & Average &  & \textit{Photo} & \textit{Art Painting} & \textit{Cartoon} & \textit{Sketch} & Average\\  \midrule
            \textit{Photo} &- &66.33 $\pm$ 1.8 &26.47$\pm$ 2.5 &32.08 $\pm$ 2.0 &41.63 $\pm$ 1.6 & \textit{Photo} &- &67.72 $\pm$ 0.7 &28.97 $\pm$ 1.0 &33.71 $\pm$ 2.6 &43.46 $\pm$ 1.1\\
            \textit{Art Painting} &96.28 $\pm$ 0.4 &- &62.54 $\pm$ 2.1 &53.19 $\pm$ 3.2 &70.67 $\pm$ 1.2 & \textit{Art Painting} & 96.62 $\pm$ 0.3 &- &65.22 $\pm$ 0.7 &55.94 $\pm$ 3.1 &72.59 $\pm$ 1.1\\
            \textit{Cartoon} &85.89 $\pm$ 1.1 &68.99 $\pm$ 1.4 &- &70.38 $\pm$ 1.7 &75.08 $\pm$ 1.0 & \textit{Cartoon} & 87.53 $\pm$ 0.8 &72.09 $\pm$ 1.2 &- &70.06 $\pm$ 1.6 &76.56 $\pm$ 0.8\\
            \textit{Sketch} &47.40 $\pm$ 3.5 &37.99 $\pm$ 1.4 &56.36 $\pm$ 3.0 &- &47.25 $\pm$ 2.9 & \textit{Sketch} & 46.07 $\pm$ 5.3 &37.17 $\pm$ 4.0 &54.03 $\pm$ 3.2 &- &45.76 $\pm$ 3.8\\ \midrule
            Average &76.52 &57.77 &48.45 &51.88 &58.66 & Average & \textbf{76.74 (+0.22\%)} &\textbf{58.99 (+1.22\%)} &\textbf{49.41 (+0.96\%)} &\textbf{53.24 (+1.36\%)} &\textbf{59.59 (+0.93\%)}\\
            \bottomrule
        \end{tabular}}
     \end{center}
\end{table*}
}
{
\setlength{\tabcolsep}{4pt}
\renewcommand{\arraystretch}{1.3} 
\begin{table*}[t]
	\begin{center}
	    \caption{Ablation study of SelfReg on PACS. 
	   \textit{Abbr.} $R_{f}$: feature-level in-batch dissimilarity loss, $R_{l}$: logit-level in-batch dissimilarity loss, Mix-up: two-domain mix-up layer, CDPL: class-specific domain perturbation layer, SWA: stochastic weights averaging, IDCL: inter-domain curriculum learning}
	    \label{tab:main_ablation}
    	\resizebox{.95\linewidth}{!}{%
    	\begin{tabular}{@{}lccccccccccc@{}} \toprule
    	    \multirow{2}{*}{Model} & \multicolumn{6}{c}{Components}& \multicolumn{4}{c}{Test Domain} & \multirow{2}{*}{Average} \\\cmidrule{2-11}
            &$\mathcal{L}_\textnormal{logit}$ &$\mathcal{L}_\textnormal{feature}$ &Mixup &CDPL  &SWA &IDCL &\textit{Photo} & \textit{Art Painting}  & \textit{Cartoon}  & \textit{Sketch}  & \\ \midrule
            A. SelfReg (Ours) &\checkmark &\checkmark &\checkmark &\checkmark &\checkmark &\checkmark &\textbf{96.22 $\pm$ 0.3} &\textbf{82.34 $\pm$ 0.5} &\textbf{78.43 $\pm$ 0.7} &\textbf{77.47 $\pm$ 0.8} &\textbf{83.62 $\pm$ 0.3} \\\midrule
            B. A w/o IDCL &\checkmark &\checkmark &\checkmark &\checkmark &\checkmark & &96.09 $\pm$ 0.3 &81.89 $\pm$ 0.6 &78.03 $\pm$ 0.4 &77.21 $\pm$ 1.1 &83.30 $\pm$ 0.3\\ 
            C. B w/o SWA &\checkmark   &\checkmark &\checkmark &\checkmark & & &96.10 $\pm$ 0.5 &81.43 $\pm$ 1.0 &77.86 $\pm$ 1.0 &76.81 $\pm$ 1.2 &83.05 $\pm$ 0.5\\
            D. C w/o CDPL &\checkmark  &\checkmark &\checkmark & & & &96.04 $\pm$ 0.4 &81.66 $\pm$ 1.3 &77.48 $\pm$ 1.2 &76.16 $\pm$ 1.3 &82.84 $\pm$ 0.6 \\
            E. D w/o Mixup &\checkmark &\checkmark & & & & &96.05 $\pm$ 0.3 &81.77 $\pm$ 1.1 &77.45 $\pm$ 1.1 &75.74 $\pm$ 1.6 &82.75 $\pm$ 0.7  \\
            F. E w/o $\mathcal{L}_\textnormal{feature}$ &\checkmark & & & & & &96.19 $\pm$ 0.3 &81.59 $\pm$ 1.2 &76.98 $\pm$ 1.3 &75.71 $\pm$ 1.3 &82.62 $\pm$ 0.5 \\\midrule
            G. F w/o $\mathcal{L}_\textnormal{logit}$ (baseline) & & & & & & &95.66 $\pm$ 0.4 &79.89 $\pm$ 1.3 &75.61 $\pm$ 1.5 &73.33 $\pm$ 2.8 &81.12 $\pm$ 0.8 \\\bottomrule
        \end{tabular}}
     \end{center}
\end{table*}
}
\subsection{Single-source Domain Generalization}\label{sec:one_source}
We also evaluate our model in an extreme case for the domain generalization task. We train our model with examples from a single source domain (not multiple source domains as we see in a previous experimental setting), and then we evaluate with examples from other remaining target domains. As shown in Table~\ref{tab:one_source_domain}, we report scores for all source-target combinations, i.e. rows and columns for source and target domains, respectively. As a baseline, we compare ours with those of RSC~\cite{huangRSC2020} evaluated in the same setting (compare scores in left and right tables). We also report their differences in the last row (`$+$' indicates that ours performs better). We observe in Table~\ref{tab:one_source_domain} that ours generally outperform alternative, where the average accuracy is improved by 0.93\%.

{
\setlength{\tabcolsep}{4pt}
\renewcommand{\arraystretch}{1.3} 
\begin{table*}[t]
	\begin{center}
	    \caption{Average out-of-distribution test accuracies on the DomainBed setting. Here we compare 14 domain generalization algorithms in the exact same conditions. Note that we train domain validation set as a model selection method. $^{\dagger}$: Ours does not use IDCL and SWA techniques due to implementational inflexibility on the DomainBed environment. \textit{Abbr.} $D$: learning domain-invariant features by matching distributions across different domains, $A$: Adversarial learning strategy, $M$: inter-domain mix-up, $C$: contrastive learning, $U$: unsupervised domain adaptation, which is originally designed to take examples from the target domain during training.}
	    \label{tab:domainbed}
    	\resizebox{\linewidth}{!}{%
    	\begin{tabular}{@{}lccccccccccccc@{}} \toprule
    	    Model & $D$ & $A$ & $M$ & $C$ & $U$ & CMNIST~\cite{arjovsky2019invariant} & RMNIST~\cite{ghifary2015domain} & VLCS~\cite{fang2013unbiased} & PACS~\cite{Li2017dg} & OfficeHome~\cite{venkateswara2017deep} & TerraIncognita~\cite{beery2018recognition} & DomainNet~\cite{peng2019moment}& Average             \\\midrule
            
            CORAL~\cite{sun2016deep} & \checkmark & & & & \checkmark
            & 51.5 $\pm$ 0.1 & 98.0 $\pm$ 0.1 & 78.8 $\pm$ 0.6 & 86.2 $\pm$ 0.3 & 68.7 $\pm$ 0.3 & 47.6 $\pm$ 1.0  & 41.5 $\pm$ 0.1 & 67.5 \\\midrule
            
            SelfReg (ours)$^{\dagger}$ & \checkmark &   & \checkmark & \checkmark & 
            & 52.1 $\pm$ 0.2 & 98.0 $\pm$ 0.1 & 77.8 $\pm$ 0.9 & 85.6 $\pm$ 0.4 & 67.9 $\pm$ 0.7 & 47.0 $\pm$ 0.3 & 42.8 $\pm$ 0.0  & 67.3\\\midrule
            
            SagNet~\cite{nam2019reducing} & \checkmark & \checkmark &  & & 
            & 51.7 $\pm$ 0.0 & 98.0 $\pm$ 0.0 & 77.8 $\pm$ 0.5 & 86.3 $\pm$ 0.2 & 68.1 $\pm$ 0.1 & 48.6 $\pm$ 1.0  & 40.3 $\pm$ 0.1 & 67.2 \\
            
            Mixup~\cite{yan2020improve} & & & \checkmark & & 
            & 52.1 $\pm$ 0.2 & 98.0 $\pm$ 0.1 & 77.4 $\pm$ 0.6 & 84.6 $\pm$ 0.6 & 68.1 $\pm$ 0.3 & 47.9 $\pm$ 0.8  & 39.2 $\pm$ 0.1 & 66.7 \\

            MLDG~\cite{li2018learning} & & &  & & 
            & 51.5 $\pm$ 0.1 & 97.9 $\pm$ 0.0 & 77.2 $\pm$ 0.4 & 84.9 $\pm$ 1.0 & 66.8 $\pm$ 0.6 & 47.7 $\pm$ 0.9  & 41.2 $\pm$ 0.1 & 66.7 \\
            
    	    ERM~\cite{vapnik1999overview} & & &  & & 
    	    & 51.5 $\pm$ 0.1 & 98.0 $\pm$ 0.0 & 77.5 $\pm$ 0.4 & 85.5 $\pm$ 0.2 & 66.5 $\pm$ 0.3 & 46.1 $\pm$ 1.8  & 40.9 $\pm$ 0.1 & 66.6 \\
            
            MTL~\cite{blanchard2017domain} & & & & & 
            & 51.4 $\pm$ 0.1 & 97.9 $\pm$ 0.0 & 77.2 $\pm$ 0.4 & 84.6 $\pm$ 0.5 & 66.4 $\pm$ 0.5 & 45.6 $\pm$ 1.2  & 40.6 $\pm$ 0.1 & 66.2 \\
            
            RSC~\cite{huangRSC2020} & & &  & & 
            & 51.7 $\pm$ 0.2 & 97.6 $\pm$ 0.1 & 77.1 $\pm$ 0.5 & 85.2 $\pm$ 0.9 & 65.5 $\pm$ 0.9 & 46.6 $\pm$ 1.0  & 38.9 $\pm$ 0.5 & 66.1 \\ 
            
            ARM~\cite{zhang2020adaptive} & & &  & & 
            & 56.2 $\pm$ 0.2 & 98.2 $\pm$ 0.1 & 77.6 $\pm$ 0.3 & 85.1 $\pm$ 0.4 & 64.8 $\pm$ 0.3 & 45.5 $\pm$ 0.3  & 35.5 $\pm$ 0.2 & 66.1 \\
            
            DANN~\cite{ganin2016domain} & \checkmark & \checkmark&  & & \checkmark
            & 51.5 $\pm$ 0.3 & 97.8 $\pm$ 0.1 & 78.6 $\pm$ 0.4 & 83.6 $\pm$ 0.4 & 65.9 $\pm$ 0.6 & 46.7 $\pm$ 0.5  & 38.3 $\pm$ 0.1 & 66.1 \\
            
            VREx~\cite{krueger2020out} & \checkmark & &  & & 
            & 51.8 $\pm$ 0.1 & 97.9 $\pm$ 0.1 & 78.3 $\pm$ 0.2 & 84.9 $\pm$ 0.6 & 66.4 $\pm$ 0.6 & 46.4 $\pm$ 0.6  & 33.6 $\pm$ 2.9 & 65.6 \\
            
            CDANN~\cite{li2018deep} & \checkmark & \checkmark&  & & 
            & 51.7 $\pm$ 0.1 & 97.9 $\pm$ 0.1 & 77.5 $\pm$ 0.1 & 82.6 $\pm$ 0.9 & 65.8 $\pm$ 1.3 & 45.8 $\pm$ 1.6  & 38.3 $\pm$ 0.3 & 65.6 \\
            
            IRM~\cite{arjovsky2019invariant} & & &  & & 
            & 52.0 $\pm$ 0.1 & 97.7 $\pm$ 0.1 & 78.5 $\pm$ 0.5 & 83.5 $\pm$ 0.8 & 64.3 $\pm$ 2.2 & 47.6 $\pm$ 0.8  & 33.9 $\pm$ 2.8 & 65.4 \\
            
            GroupDRO~\cite{sagawa2019distributionally} & \checkmark & &  & & 
            & 52.1 $\pm$ 0.0 & 98.0 $\pm$ 0.0 & 76.7 $\pm$ 0.6 & 84.4 $\pm$ 0.8 & 66.0 $\pm$ 0.7 & 43.2 $\pm$ 1.1  & 33.3 $\pm$ 0.2 & 64.8 \\
            
            MMD~\cite{li2018domain} & \checkmark & &  & & 
            & 51.5 $\pm$ 0.2 & 97.9 $\pm$ 0.0 & 77.5 $\pm$ 0.9 & 84.6 $\pm$ 0.5 & 66.3 $\pm$ 0.1 & 42.2 $\pm$ 1.6  & 23.4 $\pm$ 9.5 & 63.3 \\
            
            \bottomrule
        \end{tabular}}
     \end{center}
\end{table*}
}
\subsection{Ablation Study}\label{ss:ablation_study}
In Table~\ref{tab:main_ablation}, we compare variants of our model by removing each component: i.e. (i) feature-level in-batch dissimilarity regularization, (ii) logit-level in-batch dissimilarity regularization, (iii) a two-domain mix-up layer, (iv) a class-specific domain perturbation layer (CDPL), (v) stochastic weights averaging (SWA), and (vi) inter-domain curriculum learning (IDCL). 

\myparagraph{Effect of Inter-domain Curriculum Learning (IDCL)}
We observe in Table~\ref{tab:main_ablation} that applying our inter-domain curriculum learning (IDCL) provides the recognition accuracy (compare model A vs. B). Scores are generally improved in all target domains, i.e. the average accuracy is improved by 0.32\%. 

\myparagraph{Effect of Stochastic Weights Averaging (SWA)}
As shown in Table~\ref{tab:main_ablation}, the use of stochastic weight average technique further provides better performance (compare model B vs. C) in all target domains, i.e. the average accuracy is improved by 0.25\%. This is probably due to SWA provides the flatness in loss surface by ensembling domain-specific models, which generally have multiple local-minima during the training procedure.

\myparagraph{Effect of Mixup and CDPL}
As shown in Table~\ref{tab:main_ablation}, we observe that both CDPL and Mixup components contribute to improve the overall performance (compare Model C vs. D for CDPL, and Model D vs. E for Mixup). Such improvement is more noticeable for the \textit{Sketch} domain, which may support that CDPL reinforces the overall effect of mixup and makes DG performance more robust for target domains that are significantly distanced from their source domains.

\myparagraph{Feature and Logit-level Contrastive Losses}
Model $F$, as defined as the baseline model (Model $G$) plus $R_{l}$, had an average performance improvement of 1.50\%. Accuracy improved and variance decreased across all of the domains. Therefore, regularization to minimize the logit vector-wise distance on positive pairs appears effective in extracting domain invariant features. Furthermore, Model $E$, which adds $R_{f}$ and $R_{l}$ to the baseline model, exhibited even greater performance increase. Minimizing feature distances of positive pairs as well as logit distances, was observed to be effective in improving DG performance.

\section{Experiments on DomainBed}\label{sec:domainbed}
We further conduct experiments using DomainBed~\cite{gulrajani2020search}, which is a unified testbed useful for evaluating domain generalization algorithms. This testbed currently provides seven multi-domain datasets (i.e. ColoredMNIST~\cite{arjovsky2019invariant}, RotatedMNIST~\cite{ghifary2015domain}, VLCS~\cite{fang2013unbiased}, PACS~\cite{Li2017dg}, OfficeHome~\cite{venkateswara2017deep}, and TerraIncognita~\cite{beery2018recognition}, DomainNet~\cite{peng2019moment}) and provides benchmarks results of 14 baseline approaches (i.e. ERM~\cite{vapnik1999overview}, IRM~\cite{arjovsky2019invariant}, GroupDRO~\cite{sagawa2019distributionally}, Mixup~\cite{yan2020improve}, MLDG~\cite{li2018learning}, CORAL~\cite{sun2016deep}, MMD~\cite{li2018domain}, DANN~\cite{ganin2016domain}, CDANN~\cite{li2018deep}, MTL~\cite{blanchard2017domain}, SagNet~\cite{nam2019reducing}, ARM~\cite{zhang2020adaptive}, VREx~\cite{krueger2020out}, RSC~\cite{huangRSC2020}). 

As shown in Table~\ref{tab:domainbed}, we also report scores for our model evaluated in the setting of DomainBed. We observe in Table~\ref{tab:domainbed} that ours generally shows matched or better performance against alternative state-of-the-art methods, where ours is ranked 2nd places in terms of average of all seven benchmarks. Note that ours does not use IDCL and SWA (\ref{ss:swa}) techniques, which we confirmed that further improvements are highly achievable combined with these techniques. We provide more detailed scores for each domain in the appendix. Note that DANN~\cite{ganin2016domain} and CORAL~\cite{sun2016deep} are designed to take examples from the target domain during training -- i.e. CORAL~\cite{sun2016deep} is trained to minimize the distance between covariances of the source and target features. Note also that some studies~\cite{nam2019reducing, ganin2016domain, li2018deep} use the adversarial learning setting to obtain an unknown domain-invariant feature by fitting implicit generative models, such as GAN (generative adversarial networks). Though GAN is a powerful framework, the alternating gradient updates procedure is often highly unstable and often results in mode collapse~\cite{kodali2017convergence}.

\section{Conclusion}
In this paper, we proposed SelfReg, a new regularization method for domain generalization that leverages a self-supervised contrastive regularization loss with only positive data pairs, mitigating  problems caused by negative pair sampling.  Our experiments on PACS dataset and DomainBed benchmarks show that our model matches or outperforms prior work under the standard domain generalization evaluation setting.  In future work, it would be interesting to extend SelfReg with the siamese network, enabling the model to choose better positive data pairs.

{
\bibliographystyle{ieee_fullname}
\bibliography{main.bib}

\begin{thebibliography}{10}\itemsep=-1pt

\bibitem{arjovsky2019invariant}
Martin Arjovsky, L{\'e}on Bottou, Ishaan Gulrajani, and David Lopez-Paz.
\newblock Invariant risk minimization.
\newblock {\em arXiv preprint arXiv:1907.02893}, 2019.

\bibitem{beery2018recognition}
Sara Beery, Grant Van~Horn, and Pietro Perona.
\newblock Recognition in terra incognita.
\newblock In {\em Proceedings of the European Conference on Computer Vision
  (ECCV)}, pages 456--473, 2018.

\bibitem{blanchard2017domain}
Gilles Blanchard, Aniket~Anand Deshmukh, Urun Dogan, Gyemin Lee, and Clayton
  Scott.
\newblock Domain generalization by marginal transfer learning.
\newblock {\em arXiv preprint arXiv:1711.07910}, 2017.

\bibitem{blanchard2011generalizing}
Gilles Blanchard, Gyemin Lee, and Clayton Scott.
\newblock Generalizing from several related classification tasks to a new
  unlabeled sample.
\newblock {\em Advances in neural information processing systems},
  24:2178--2186, 2011.

\bibitem{carlucci2019domain}
Fabio~M Carlucci, Antonio D'Innocente, Silvia Bucci, Barbara Caputo, and
  Tatiana Tommasi.
\newblock Domain generalization by solving jigsaw puzzles.
\newblock In {\em Proceedings of the IEEE Conference on Computer Vision and
  Pattern Recognition}, pages 2229--2238, 2019.

\bibitem{cha2021domain}
Junbum Cha, Hancheol Cho, Kyungjae Lee, Seunghyun Park, Yunsung Lee, and
  Sungrae Park.
\newblock Domain generalization needs stochastic weight averaging for
  robustness on domain shifts, 2021.

\bibitem{chen2020simple}
Ting Chen, Simon Kornblith, Mohammad Norouzi, and Geoffrey Hinton.
\newblock A simple framework for contrastive learning of visual
  representations.
\newblock In {\em International conference on machine learning}, pages
  1597--1607. PMLR, 2020.

\bibitem{chen2020exploring}
Xinlei Chen and Kaiming He.
\newblock Exploring simple siamese representation learning.
\newblock {\em arXiv preprint arXiv:2011.10566}, 2020.

\bibitem{deng2009imagenet}
Jia Deng, Wei Dong, Richard Socher, Li-Jia Li, Kai Li, and Li Fei-Fei.
\newblock Imagenet: A large-scale hierarchical image database.
\newblock In {\em 2009 IEEE conference on computer vision and pattern
  recognition}, pages 248--255. Ieee, 2009.

\bibitem{donahuedeep}
Jeff Donahue, Yangqing Jia, Oriol Vinyals, Judy Hoffman, Ning Zhang, Eric
  Tzeng, and Trevor Darrell.
\newblock A deep convolutional activation feature for generic visual
  recognition.
\newblock {\em UC Berkeley \& ICSI, Berkeley, CA, USA}.

\bibitem{dou2019domain}
Qi Dou, Daniel~C Castro, Konstantinos Kamnitsas, and Ben Glocker.
\newblock Domain generalization via model-agnostic learning of semantic
  features.
\newblock {\em arXiv preprint arXiv:1910.13580}, 2019.

\bibitem{d2018domain}
Antonio D’Innocente and Barbara Caputo.
\newblock Domain generalization with domain-specific aggregation modules.
\newblock In {\em German Conference on Pattern Recognition}, pages 187--198.
  Springer, 2018.

\bibitem{fang2013unbiased}
Chen Fang, Ye Xu, and Daniel~N Rockmore.
\newblock Unbiased metric learning: On the utilization of multiple datasets and
  web images for softening bias.
\newblock In {\em Proceedings of the IEEE International Conference on Computer
  Vision}, pages 1657--1664, 2013.

\bibitem{finn2017model}
Chelsea Finn, Pieter Abbeel, and Sergey Levine.
\newblock Model-agnostic meta-learning for fast adaptation of deep networks.
\newblock In {\em International Conference on Machine Learning}, pages
  1126--1135. PMLR, 2017.

\bibitem{ganin2016domain}
Yaroslav Ganin, Evgeniya Ustinova, Hana Ajakan, Pascal Germain, Hugo
  Larochelle, Fran{\c{c}}ois Laviolette, Mario Marchand, and Victor Lempitsky.
\newblock Domain-adversarial training of neural networks.
\newblock {\em The journal of machine learning research}, 17(1):2096--2030,
  2016.

\bibitem{ghifary2015domain}
Muhammad Ghifary, W~Bastiaan Kleijn, Mengjie Zhang, and David Balduzzi.
\newblock Domain generalization for object recognition with multi-task
  autoencoders.
\newblock In {\em Proceedings of the IEEE international conference on computer
  vision}, pages 2551--2559, 2015.

\bibitem{grill2020bootstrap}
Jean-Bastien Grill, Florian Strub, Florent Altch{\'e}, Corentin Tallec,
  Pierre~H Richemond, Elena Buchatskaya, Carl Doersch, Bernardo~Avila Pires,
  Zhaohan~Daniel Guo, Mohammad~Gheshlaghi Azar, et~al.
\newblock Bootstrap your own latent: A new approach to self-supervised
  learning.
\newblock {\em arXiv preprint arXiv:2006.07733}, 2020.

\bibitem{gulrajani2020search}
Ishaan Gulrajani and David Lopez-Paz.
\newblock In search of lost domain generalization.
\newblock {\em arXiv preprint arXiv:2007.01434}, 2020.

\bibitem{he2019asymmetric}
Haowei He, Gao Huang, and Yang Yuan.
\newblock Asymmetric valleys: Beyond sharp and flat local minima.
\newblock {\em arXiv preprint arXiv:1902.00744}, 2019.

\bibitem{he2020momentum}
Kaiming He, Haoqi Fan, Yuxin Wu, Saining Xie, and Ross Girshick.
\newblock Momentum contrast for unsupervised visual representation learning.
\newblock In {\em Proceedings of the IEEE/CVF Conference on Computer Vision and
  Pattern Recognition}, pages 9729--9738, 2020.

\bibitem{he2016deep}
Kaiming He, Xiangyu Zhang, Shaoqing Ren, and Jian Sun.
\newblock Deep residual learning for image recognition.
\newblock In {\em Proceedings of the IEEE conference on computer vision and
  pattern recognition}, pages 770--778, 2016.

\bibitem{huangRSC2020}
Zeyi Huang, Haohan Wang, Eric~P. Xing, and Dong Huang.
\newblock Self-challenging improves cross-domain generalization.
\newblock In {\em ECCV}, 2020.

\bibitem{izmailov2018averaging}
Pavel Izmailov, Dmitrii Podoprikhin, Timur Garipov, Dmitry Vetrov, and
  Andrew~Gordon Wilson.
\newblock Averaging weights leads to wider optima and better generalization.
\newblock {\em arXiv preprint arXiv:1803.05407}, 2018.

\bibitem{kodali2017convergence}
Naveen Kodali, Jacob Abernethy, James Hays, and Zsolt Kira.
\newblock On convergence and stability of gans.
\newblock {\em arXiv preprint arXiv:1705.07215}, 2017.

\bibitem{krueger2020out}
David Krueger, Ethan Caballero, Joern-Henrik Jacobsen, Amy Zhang, Jonathan
  Binas, Dinghuai Zhang, Remi~Le Priol, and Aaron Courville.
\newblock Out-of-distribution generalization via risk extrapolation (rex).
\newblock {\em arXiv preprint arXiv:2003.00688}, 2020.

\bibitem{Li2017dg}
Da Li, Yongxin Yang, Yi-Zhe Song, and Timothy Hospedales.
\newblock Deeper, broader and artier domain generalization.
\newblock In {\em International Conference on Computer Vision}, 2017.

\bibitem{li2018learning}
Da Li, Yongxin Yang, Yi-Zhe Song, and Timothy Hospedales.
\newblock Learning to generalize: Meta-learning for domain generalization.
\newblock In {\em Proceedings of the AAAI Conference on Artificial
  Intelligence}, volume~32, 2018.

\bibitem{li2018domain}
Haoliang Li, Sinno~Jialin Pan, Shiqi Wang, and Alex~C Kot.
\newblock Domain generalization with adversarial feature learning.
\newblock In {\em Proceedings of the IEEE Conference on Computer Vision and
  Pattern Recognition}, pages 5400--5409, 2018.

\bibitem{li2018deep}
Ya Li, Xinmei Tian, Mingming Gong, Yajing Liu, Tongliang Liu, Kun Zhang, and
  Dacheng Tao.
\newblock Deep domain generalization via conditional invariant adversarial
  networks.
\newblock In {\em Proceedings of the European Conference on Computer Vision
  (ECCV)}, pages 624--639, 2018.

\bibitem{motiian2017unified}
Saeid Motiian, Marco Piccirilli, Donald~A Adjeroh, and Gianfranco Doretto.
\newblock Unified deep supervised domain adaptation and generalization.
\newblock In {\em Proceedings of the IEEE international conference on computer
  vision}, pages 5715--5725, 2017.

\bibitem{muandet2013domain}
Krikamol Muandet, David Balduzzi, and Bernhard Sch{\"o}lkopf.
\newblock Domain generalization via invariant feature representation.
\newblock In {\em International Conference on Machine Learning}, pages 10--18.
  PMLR, 2013.

\bibitem{nam2019reducing}
Hyeonseob Nam, HyunJae Lee, Jongchan Park, Wonjun Yoon, and Donggeun Yoo.
\newblock Reducing domain gap via style-agnostic networks.
\newblock {\em arXiv preprint arXiv:1910.11645}, 2019.

\bibitem{oord2018representation}
Aaron van~den Oord, Yazhe Li, and Oriol Vinyals.
\newblock Representation learning with contrastive predictive coding.
\newblock {\em arXiv preprint arXiv:1807.03748}, 2018.

\bibitem{peng2019moment}
Xingchao Peng, Qinxun Bai, Xide Xia, Zijun Huang, Kate Saenko, and Bo Wang.
\newblock Moment matching for multi-source domain adaptation.
\newblock In {\em Proceedings of the IEEE International Conference on Computer
  Vision}, pages 1406--1415, 2019.

\bibitem{peng2015learning}
Xingchao Peng, Baochen Sun, Karim Ali, and Kate Saenko.
\newblock Learning deep object detectors from 3d models.
\newblock In {\em Proceedings of the IEEE International Conference on Computer
  Vision}, pages 1278--1286, 2015.

\bibitem{sagawa2019distributionally}
Shiori Sagawa, Pang~Wei Koh, Tatsunori~B Hashimoto, and Percy Liang.
\newblock Distributionally robust neural networks for group shifts: On the
  importance of regularization for worst-case generalization.
\newblock {\em arXiv preprint arXiv:1911.08731}, 2019.

\bibitem{selvaraju2017grad}
Ramprasaath~R Selvaraju, Michael Cogswell, Abhishek Das, Ramakrishna Vedantam,
  Devi Parikh, and Dhruv Batra.
\newblock Grad-cam: Visual explanations from deep networks via gradient-based
  localization.
\newblock In {\em Proceedings of the IEEE international conference on computer
  vision}, pages 618--626, 2017.

\bibitem{sun2016deep}
Baochen Sun and Kate Saenko.
\newblock Deep coral: Correlation alignment for deep domain adaptation.
\newblock In {\em European conference on computer vision}, pages 443--450.
  Springer, 2016.

\bibitem{van2008visualizing}
Laurens Van~der Maaten and Geoffrey Hinton.
\newblock Visualizing data using t-sne.
\newblock {\em Journal of machine learning research}, 9(11), 2008.

\bibitem{vapnik1998statistical}
V Vapnik.
\newblock Statistical learning theory new york.
\newblock {\em NY: Wiley}, 1998.

\bibitem{vapnik1999overview}
Vladimir~N Vapnik.
\newblock An overview of statistical learning theory.
\newblock {\em IEEE transactions on neural networks}, 10(5):988--999, 1999.

\bibitem{venkateswara2017deep}
Hemanth Venkateswara, Jose Eusebio, Shayok Chakraborty, and Sethuraman
  Panchanathan.
\newblock Deep hashing network for unsupervised domain adaptation.
\newblock In {\em Proceedings of the IEEE conference on computer vision and
  pattern recognition}, pages 5018--5027, 2017.

\bibitem{wang2020heterogeneous}
Yufei Wang, Haoliang Li, and Alex~C Kot.
\newblock Heterogeneous domain generalization via domain mixup.
\newblock In {\em ICASSP 2020-2020 IEEE International Conference on Acoustics,
  Speech and Signal Processing (ICASSP)}, pages 3622--3626. IEEE, 2020.

\bibitem{xu2020adversarial}
Minghao Xu, Jian Zhang, Bingbing Ni, Teng Li, Chengjie Wang, Qi Tian, and
  Wenjun Zhang.
\newblock Adversarial domain adaptation with domain mixup.
\newblock In {\em Proceedings of the AAAI Conference on Artificial
  Intelligence}, volume~34, pages 6502--6509, 2020.

\bibitem{yan2020improve}
Shen Yan, Huan Song, Nanxiang Li, Lincan Zou, and Liu Ren.
\newblock Improve unsupervised domain adaptation with mixup training.
\newblock {\em arXiv preprint arXiv:2001.00677}, 2020.

\bibitem{zhang2017mixup}
Hongyi Zhang, Moustapha Cisse, Yann~N Dauphin, and David Lopez-Paz.
\newblock mixup: Beyond empirical risk minimization.
\newblock {\em arXiv preprint arXiv:1710.09412}, 2017.

\bibitem{zhang2020adaptive}
Marvin Zhang, Henrik Marklund, Abhishek Gupta, Sergey Levine, and Chelsea Finn.
\newblock Adaptive risk minimization: A meta-learning approach for tackling
  group shift.
\newblock {\em arXiv preprint arXiv:2007.02931}, 2020.

\bibitem{zhao2019learning}
Han Zhao, Remi~Tachet Des~Combes, Kun Zhang, and Geoffrey Gordon.
\newblock On learning invariant representations for domain adaptation.
\newblock In {\em International Conference on Machine Learning}, pages
  7523--7532. PMLR, 2019.

\end{thebibliography}
}
\newpage

\section*{Appendix}
In Table~\ref{sup:db_cmnist}-\ref{sup:db_domainnet}, SelfReg (ours)$^{\dagger}$ does not include Inter-domain Curriculum Learning (IDCL) and SelfReg with stochastic weight averaging (SWA)~\cite{izmailov2018averaging} techniques. However, note that Table~\ref{sup:db_pacs} provides the performance of our SelfReg with SWA technique also. Since DomainBed is supposed to be evaluated every $N$ steps, we needed to modify the code to apply the SWA technique. We modified the code to evaluate model on the test set after completing 5000 steps learning with SWA techniques. We used "--single\_test\_envs" option because the required amount of computation for the cross-validation model selection method was too much for us.

\begin{table*}[ht]
\begin{center}
\caption{Detailed scores on ColoredMNIST~\cite{arjovsky2019invariant} in DomainBed~\cite{gulrajani2020search}.}\label{sup:db_cmnist}
\adjustbox{max width=\textwidth}{%
\begin{tabular}{lcccc}
\toprule
\rowcolor{Gray}
\multicolumn{5}{c}{Model selection: training-domain validation set} \\  
\midrule
Algorithm   & +90\%       & +80\%       & -90\%         & Avg         \\ 
\midrule
SelfReg (ours)$^{\dagger}$       & 72.2 $\pm$ 0.5	    &73.7 $\pm$ 0.2	       & 10.5 $\pm$ 0.3 	& 52.1 $\pm$ 0.2    \\
ERM~\cite{vapnik1999overview}                  & 71.7 $\pm$ 0.1       & 72.9 $\pm$ 0.2       & 10.0 $\pm$ 0.1       & 51.5 \\
\toprule
\rowcolor{Gray}
\multicolumn{5}{c}{Model selection: test-domain validation set (oracle)} \\ 
\midrule
Algorithm   & +90\%       & +80\%       & -90\%         & Avg         \\ 
\midrule
SelfReg (ours)$^{\dagger}$       & 71.3 $\pm$ 0.4	&73.4 $\pm$ 0.2	 &29.3 $\pm$ 2.1 	&58.0 $\pm$ 0.7    \\ 
ERM~\cite{vapnik1999overview}                  & 71.8 $\pm$ 0.4       & 72.9 $\pm$ 0.1       & 28.7 $\pm$ 0.5       & 57.8 \\
\bottomrule
\end{tabular}}
\end{center}
\end{table*}
\begin{table*}[ht]
\begin{center}
\caption{Detailed scores on RotatedMNIST~\cite{ghifary2015domain} in DomainBed~\cite{gulrajani2020search}.}\label{sup:db_rmnist}
\adjustbox{max width=\textwidth}{%
\begin{tabular}{lccccccc}
\toprule
\rowcolor{Gray}
\multicolumn{8}{c}{Model selection: training-domain validation set} \\  
\midrule
Algorithm   &0           &15          &30          &45          &60          &75          & Avg         \\
\midrule
SelfReg (ours)$^{\dagger}$       & 95.7 $\pm$ 0.3	& 99.0 $\pm$ 0.1	& 98.9 $\pm$ 0.1	& 99.0 $\pm$ 0.1	& 98.9 $\pm$ 0.1	& 96.6 $\pm$ 0.1	& 98.0 $\pm$ 0.2    \\
ERM~\cite{vapnik1999overview}                  & 95.9 $\pm$ 0.1       & 98.9 $\pm$ 0.0       & 98.8 $\pm$ 0.0       & 98.9 $\pm$ 0.0       & 98.9 $\pm$ 0.0       & 96.4 $\pm$ 0.0       & 98.0 \\

\toprule
\rowcolor{Gray}
\multicolumn{8}{c}{Model selection: test-domain validation set (oracle)} \\ 
\midrule
Algorithm   &0           &15          &30          &45          &60          &75          & Avg         \\
\midrule
SelfReg (ours)$^{\dagger}$       & 96.0 $\pm$ 0.3	& 98.9 $\pm$ 0.1	& 98.9 $\pm$ 0.1	& 98.9 $\pm$ 0.1	& 98.9 $\pm$ 0.1	& 96.8 $\pm$ 0.1	& 98.1 $\pm$ 0.7    \\ 
ERM~\cite{vapnik1999overview}                  & 95.3 $\pm$ 0.2       & 98.7 $\pm$ 0.1       & 98.9 $\pm$ 0.1       & 98.7 $\pm$ 0.2       & 98.9 $\pm$ 0.0       & 96.2 $\pm$ 0.2       & 97.8 \\
\bottomrule
\end{tabular}}
\end{center}
\end{table*}

\begin{table*}[ht]
\begin{center}
\caption{Detailed scores on VLCS~\cite{fang2013unbiased} in DomainBed~\cite{gulrajani2020search}.}\label{sup:db_vlcs}
\adjustbox{max width=\textwidth}{%
\begin{tabular}{lccccc}
\toprule
\rowcolor{Gray}
\multicolumn{6}{c}{Model selection: training-domain validation set} \\  \midrule
Algorithm   & C       & L       & S         & V     & Avg\\ \midrule
SelfReg (ours)$^{\dagger}$       & 96.7 $\pm$ 0.4	    & 65.2 $\pm$ 1.2	       & 73.1 $\pm$ 1.3 	& 76.2 $\pm$ 0.7    & 77.8 $\pm$ 0.9 \\
ERM~\cite{vapnik1999overview}       & 97.7 $\pm$ 0.4       & 64.3 $\pm$ 0.9       & 73.4 $\pm$ 0.5       & 74.6 $\pm$ 1.3       & 77.5     \\ 
\toprule
\rowcolor{Gray}
\multicolumn{6}{c}{Model selection: test-domain validation set (oracle)} \\ \midrule
Algorithm   & C       & L       & S         & V         & Avg\\ \midrule
SelfReg (ours)$^{\dagger}$       & 97.9 $\pm$ 0.4	    & 66.7 $\pm$ 0.1	       & 73.5 $\pm$ 0.7 	& 74.7 $\pm$ 0.7    & 78.2 $\pm$ 0.1 \\ 
ERM~\cite{vapnik1999overview}       & 97.6 $\pm$ 0.3       & 67.9 $\pm$ 0.7       & 70.9 $\pm$ 0.2       & 74.0 $\pm$ 0.6       & 77.6 \\ \bottomrule
\end{tabular}}
\end{center}
\end{table*}

\begin{table*}[ht]
\begin{center}
\caption{Detailed scores on PACS~\cite{Li2017dg} in DomainBed~\cite{gulrajani2020search}. Note that we provide the performance of our SelfReg applied with SWA~\cite{izmailov2018averaging} technique also. }\label{sup:db_pacs}
\adjustbox{max width=\textwidth}{%
\begin{tabular}{lccccc}
\toprule
\rowcolor{Gray}
\multicolumn{6}{c}{Model selection: training-domain validation set} \\  \midrule
Algorithm   & A       & C       & P         & S     & Avg\\ \midrule
SelfReg with SWA (ours)     & 85.9 $\pm$ 0.6  & 81.9 $\pm$ 0.4   & 96.8 $\pm$ 0.1   & 81.4 $\pm$ 0.6   & 86.5 $\pm$ 0.3        \\ \midrule
SagNet~\cite{nam2019reducing}              & 87.4 $\pm$ 1.0       & 80.7 $\pm$ 0.6       & 97.1 $\pm$ 0.1       & 80.0 $\pm$ 0.4       & 86.3                 \\
CORAL~\cite{sun2016deep}                & 88.3 $\pm$ 0.2       & 80.0 $\pm$ 0.5       & 97.5 $\pm$ 0.3       & 78.8 $\pm$ 1.3       & 86.2                 \\
SelfReg (ours)$^{\dagger}$       & 87.9 $\pm$ 1.0	& 79.4 $\pm$ 1.4	& 96.8 $\pm$ 0.7	& 78.3 $\pm$ 1.2	& 85.6 $\pm$ 0.4 \\
ERM~\cite{vapnik1999overview}                  & 84.7 $\pm$ 0.4       & 80.8 $\pm$ 0.6       & 97.2 $\pm$ 0.3       & 79.3 $\pm$ 1.0       & 85.5              \\
RSC~\cite{huangRSC2020}                  & 85.4 $\pm$ 0.8       & 79.7 $\pm$ 1.8       & 97.6 $\pm$ 0.3       & 78.2 $\pm$ 1.2       & 85.2                 \\
ARM~\cite{zhang2020adaptive}                  & 86.8 $\pm$ 0.6       & 76.8 $\pm$ 0.5       & 97.4 $\pm$ 0.3       & 79.3 $\pm$ 1.2       & 85.1                 \\
VREx~\cite{krueger2020out}                 & 86.0 $\pm$ 1.6       & 79.1 $\pm$ 0.6       & 96.9 $\pm$ 0.5       & 77.7 $\pm$ 1.7       & 84.9                 \\
MLDG~\cite{li2018learning}                 & 85.5 $\pm$ 1.4       & 80.1 $\pm$ 1.7       & 97.4 $\pm$ 0.3       & 76.6 $\pm$ 1.1       & 84.9                 \\
MMD~\cite{li2018domain}                  & 86.1 $\pm$ 1.4       & 79.4 $\pm$ 0.9       & 96.6 $\pm$ 0.2       & 76.5 $\pm$ 0.5       & 84.6                 \\
Mixup~\cite{yan2020improve}                & 86.1 $\pm$ 0.5       & 78.9 $\pm$ 0.8       & 97.6 $\pm$ 0.1       & 75.8 $\pm$ 1.8       & 84.6                 \\
MTL~\cite{blanchard2017domain}                  & 87.5 $\pm$ 0.8       & 77.1 $\pm$ 0.5       & 96.4 $\pm$ 0.8       & 77.3 $\pm$ 1.8       & 84.6                 \\
GroupDRO~\cite{sagawa2019distributionally}             & 83.5 $\pm$ 0.9       & 79.1 $\pm$ 0.6       & 96.7 $\pm$ 0.3       & 78.3 $\pm$ 2.0       & 84.4        \\
DANN~\cite{ganin2016domain}                 & 86.4 $\pm$ 0.8       & 77.4 $\pm$ 0.8       & 97.3 $\pm$ 0.4       & 73.5 $\pm$ 2.3       & 83.6                 \\
IRM~\cite{arjovsky2019invariant}                  & 84.8 $\pm$ 1.3       & 76.4 $\pm$ 1.1       & 96.7 $\pm$ 0.6       & 76.1 $\pm$ 1.0       & 83.5                 \\
CDANN~\cite{li2018deep}                & 84.6 $\pm$ 1.8       & 75.5 $\pm$ 0.9       & 96.8 $\pm$ 0.3       & 73.5 $\pm$ 0.6       & 82.6                 \\
\toprule
\rowcolor{Gray}
\multicolumn{6}{c}{Model selection: test-domain validation set (oracle)} \\ \midrule
Algorithm   & A       & C       & P         & S         & Avg\\ \midrule
SelfReg with SWA (ours)     & 87.5 $\pm$ 0.1     & 83.0 $\pm$ 0.1     & 97.6 $\pm$ 0.1      & 82.8 $\pm$ 0.2    & 87.7 $\pm$ 0.1     \\ \midrule
SagNet~\cite{nam2019reducing}               & 87.4 $\pm$ 0.5       & 81.2 $\pm$ 1.2       & 96.3 $\pm$ 0.8       & 80.7 $\pm$ 1.1       & 86.4                 \\
CORAL~\cite{sun2016deep}                & 86.6 $\pm$ 0.8       & 81.8 $\pm$ 0.9       & 97.1 $\pm$ 0.5       & 82.7 $\pm$ 0.6       & 87.1                 \\
SelfReg (ours)$^{\dagger}$       & 87.9 $\pm$ 0.5	    & 80.6 $\pm$ 1.1	       & 97.1 $\pm$ 0.4 	& 81.1 $\pm$ 1.3    & 86.7 $\pm$ 0.8 \\ 
ERM~\cite{vapnik1999overview}       & 86.5 $\pm$ 1.0       & 81.3 $\pm$ 0.6       & 96.2 $\pm$ 0.3       & 82.7 $\pm$ 1.1       & 86.7 \\ 
RSC~\cite{huangRSC2020}                  & 86.0 $\pm$ 0.7       & 81.8 $\pm$ 0.9       & 96.8 $\pm$ 0.7       & 80.4 $\pm$ 0.5       & 86.2                 \\
ARM~\cite{zhang2020adaptive}                  & 85.0 $\pm$ 1.2       & 81.4 $\pm$ 0.2       & 95.9 $\pm$ 0.3       & 80.9 $\pm$ 0.5       & 85.8                 \\
VREx~\cite{krueger2020out}                 & 87.8 $\pm$ 1.2       & 81.8 $\pm$ 0.7       & 97.4 $\pm$ 0.2       & 82.1 $\pm$ 0.7       & 87.2                 \\
MLDG~\cite{li2018learning}                 & 87.0 $\pm$ 1.2       & 82.5 $\pm$ 0.9       & 96.7 $\pm$ 0.3       & 81.2 $\pm$ 0.6       & 86.8                 \\
MMD~\cite{li2018domain}                  & 88.1 $\pm$ 0.8       & 82.6 $\pm$ 0.7       & 97.1 $\pm$ 0.5       & 81.2 $\pm$ 1.2       & 87.2                 \\
Mixup~\cite{yan2020improve}                & 87.5 $\pm$ 0.4       & 81.6 $\pm$ 0.7       & 97.4 $\pm$ 0.2       & 80.8 $\pm$ 0.9       & 86.8                 \\
MTL~\cite{blanchard2017domain}                  & 87.0 $\pm$ 0.2       & 82.7 $\pm$ 0.8       & 96.5 $\pm$ 0.7       & 80.5 $\pm$ 0.8       & 86.7                 \\
GroupDRO~\cite{sagawa2019distributionally}             & 87.5 $\pm$ 0.5       & 82.9 $\pm$ 0.6       & 97.1 $\pm$ 0.3       & 81.1 $\pm$ 1.2       & 87.1                 \\
DANN~\cite{ganin2016domain}                 & 87.0 $\pm$ 0.4       & 80.3 $\pm$ 0.6       & 96.8 $\pm$ 0.3       & 76.9 $\pm$ 1.1       & 85.2                 \\
IRM~\cite{arjovsky2019invariant}                  & 84.2 $\pm$ 0.9       & 79.7 $\pm$ 1.5       & 95.9 $\pm$ 0.4       & 78.3 $\pm$ 2.1       & 84.5                 \\
CDANN~\cite{li2018deep}                & 87.7 $\pm$ 0.6       & 80.7 $\pm$ 1.2       & 97.3 $\pm$ 0.4       & 77.6 $\pm$ 1.5       & 85.8                 \\
\bottomrule
\end{tabular}}
\end{center}
\end{table*}

\begin{table*}[ht]
\begin{center}
\caption{Detailed scores on OfficeHome~\cite{venkateswara2017deep} in DomainBed~\cite{gulrajani2020search}. 
}\label{sup:db_officehome}
\adjustbox{max width=\textwidth}{%
\begin{tabular}{lccccc}
\toprule
\rowcolor{Gray}
\multicolumn{6}{c}{Model selection: training-domain validation set} \\  \midrule
Algorithm   & A       & C       & P         & R     & Avg\\ \midrule
SelfReg (ours)$^{\dagger}$       & 63.6 $\pm$ 1.4	    & 53.1 $\pm$ 1.0	       & 76.9 $\pm$ 0.4 	& 78.1 $\pm$ 0.4    & 67.9 $\pm$ 0.7 \\
ERM~\cite{vapnik1999overview}       & 61.3 $\pm$ 0.7       & 52.4 $\pm$ 0.3       & 75.8 $\pm$ 0.1       & 76.6 $\pm$ 0.3       & 66.5     \\ 
\toprule
\rowcolor{Gray}
\multicolumn{6}{c}{Model selection: test-domain validation set (oracle)} \\ \midrule
Algorithm   & A       & C       & P         & R         & Avg\\ \midrule
SelfReg (ours)$^{\dagger}$       & 64.2 $\pm$ 0.6	    & 53.6 $\pm$ 0.7	       & 76.7 $\pm$ 0.3 	& 77.9 $\pm$ 0.5    & 68.1 $\pm$ 0.3 \\ 
ERM~\cite{vapnik1999overview}       & 61.7 $\pm$ 0.7       & 53.4 $\pm$ 0.3       & 74.1 $\pm$ 0.4       & 76.2 $\pm$ 0.6       & 66.4  \\ \bottomrule
\end{tabular}}
\end{center}
\end{table*}

\begin{table*}[ht]
\begin{center}
\caption{Detailed scores on TerraIncognita~\cite{beery2018recognition} in DomainBed~\cite{gulrajani2020search}.}\label{sup:db_terra}
\adjustbox{max width=\textwidth}{%
\begin{tabular}{lccccc}
\toprule
\rowcolor{Gray}
\multicolumn{6}{c}{Model selection: training-domain validation set} \\  \midrule
Algorithm   & L100       & L38       & L43         & L46     & Avg\\ \midrule
SelfReg (ours)$^{\dagger}$       & 48.8 $\pm$ 0.9	    & 41.3 $\pm$ 1.8	       & 57.3 $\pm$ 0.7 	& 40.6 $\pm$ 0.9    & 47.0 $\pm$ 0.3 \\
ERM~\cite{vapnik1999overview}       & 49.8 $\pm$ 4.4       & 42.1 $\pm$ 1.4       & 56.9 $\pm$ 1.8       & 35.7 $\pm$ 3.9       & 46.1     \\ 
\toprule
\rowcolor{Gray}
\multicolumn{6}{c}{Model selection: test-domain validation set (oracle)} \\ \midrule
Algorithm   & L100       & L38       & L43         & L46         & Avg\\ \midrule
SelfReg (ours)$^{\dagger}$       & 60.0 $\pm$ 2.3	    & 48.8 $\pm$ 1.0	       & 58.6 $\pm$ 0.8 	& 44.0 $\pm$ 0.6    & 52.8 $\pm$ 0.9 \\ 
ERM~\cite{vapnik1999overview}       & 59.4 $\pm$ 0.9       & 49.3 $\pm$ 0.6       & 60.1 $\pm$ 1.1       & 43.2 $\pm$ 0.5       & 53.0  \\ \bottomrule
\end{tabular}}
\end{center}
\end{table*}

\begin{table*}[ht]
\begin{center}
\caption{Detailed scores on DomainNet~\cite{peng2019moment} in DomainBed~\cite{gulrajani2020search}.
Note that SelfReg$^{\dagger}$ achived the-state-of-the-art.}\label{sup:db_domainnet}
\adjustbox{max width=\textwidth}{%
\begin{tabular}{lccccccc}
\toprule
\rowcolor{Gray}
\multicolumn{8}{c}{Model selection: training-domain validation set} \\  
\midrule
Algorithm   &Clip           &Info          &Paint          &Quick          &Real          &Sketch          & Avg         \\
\midrule
SelfReg (ours)$^{\dagger}$       & 60.7 $\pm$ 0.1	& 21.6 $\pm$ 0.1	& 49.4 $\pm$ 0.2	& 12.7 $\pm$ 0.1	& 60.7 $\pm$ 0.1	& 51.7 $\pm$ 0.1	& 42.8 $\pm$ 0.0    \\
ERM~\cite{vapnik1999overview}                  & 58.1 $\pm$ 0.3       & 18.8 $\pm$ 0.3       & 46.7 $\pm$ 0.3       & 12.2 $\pm$ 0.4       & 59.6 $\pm$ 0.1       & 49.8 $\pm$ 0.4       & 40.9\\

\toprule
\rowcolor{Gray}
\multicolumn{8}{c}{Model selection: test-domain validation set (oracle)} \\ 
\midrule
Algorithm   &Clip           &Info          &Paint          &Quick          &Real          &Sketch          & Avg         \\
\midrule
SelfReg (ours)$^{\dagger}$       & 60.7 $\pm$ 0.1	& 21.6 $\pm$ 0.1	& 49.5 $\pm$ 0.1	& 14.2 $\pm$ 0.3	& 60.7 $\pm$ 0.1	& 51.7 $\pm$ 0.1	& 43.1 $\pm$ 0.1    \\
ERM~\cite{vapnik1999overview}                  & 58.6 $\pm$ 0.3       & 19.2 $\pm$ 0.2       & 47.0 $\pm$ 0.3       & 13.2 $\pm$ 0.2       & 59.9 $\pm$ 0.3       & 49.8 $\pm$ 0.4       & 41.3 \\
\bottomrule
\end{tabular}}
\end{center}
\end{table*}
\end{document}